\documentclass{article}

\usepackage{arxiv}

\usepackage[utf8]{inputenc} 
\usepackage[T1]{fontenc}    
\usepackage{hyperref}       
\usepackage{url}            
\usepackage{booktabs}       
\usepackage{amsfonts}       
\usepackage{nicefrac}       
\usepackage{microtype}      
\usepackage{graphicx}
\usepackage{array}
\newcolumntype{P}[1]{>{\raggedright\arraybackslash}p{#1}} 
\usepackage{tabularx}
\newcolumntype{Y}{>{\raggedright\arraybackslash}X}   
\usepackage{enumitem}          
\usepackage{float}             
\usepackage{float}   

\graphicspath{ {./images/} }

\usepackage{listings}
\title{Propaganda and Information Dissemination in the Russo-Ukrainian War
: Natural Language Processing of Russian and Western Twitter Narratives}

\author{
  Zaur Gouliev\\
  University College Dublin, School of Information \& Communication Studies\\
  \texttt{zaur.gouliev@ucdconnect.ie}  
}

\begin{document}
\maketitle
\begin{abstract}
The conflict in Ukraine has been not only characterised by military engagement but also by a significant information war, with social media platforms like X, formerly known as Twitter playing an important role in shaping public perception. This article provides an analysis of tweets from propaganda accounts and trusted accounts collected from the onset of the war, February 2022 until the middle of May 2022 with n=40,000 total tweets. We utilise natural language processing and machine learning algorithms to assess the sentiment and identify key themes, topics and narratives across the dataset with human-in-the-loop (HITL) analysis throughout. Our findings indicate distinct strategies in how information is created, spread, and targeted at different audiences by both sides. Propaganda accounts frequently employ emotionally charged language and disinformation to evoke fear and distrust, whereas other accounts, primarily Western tend to focus on factual reporting and humanitarian aspects of the conflict. Clustering analysis reveals groups of accounts with similar behaviours, which we suspect indicates the presence of coordinated efforts. This research attempts to contribute to our understanding of the dynamics of information warfare and offers techniques for future studies on social media influence in military conflicts.
\end{abstract}

\maketitle

\section{Introduction}

The conflict in Ukraine, which initially started in 2014 \cite{masters2023ukraine}  when Russia invaded the Crimean Peninsula \cite{bbc2018crimea}, and then annexed it \cite{crisisgroup2021donbas} has been marked not only by military confrontations but also by an intense information war \cite{perez2022information} that has unfolded across various digital platforms. In particular, social media platforms like Twitter, now re-branded as X, have played a pivotal role in shaping public opinion and influencing global perceptions of the conflict \cite{breve2024russiaukraine}. This phenomenon is not unique to the Ukraine conflict; rather, it represents a broader trend in modern warfare, where information dissemination and manipulation have become crucial components of state and non-state actors strategies \cite{woolley2017computational}, \cite{prier2017commanding}. The importance of social media in modern conflicts \cite{nissen2023social}, \cite{bachmann2023hybrid} is the main concern for this article, which prompts the need for a deeper analysis to understand how these platforms are used to propagate narratives, sway public sentiment, and conduct influence operations \cite{pamment2020new}. Our research question (RQ) is twofold, given the research that social media influences modern conflicts; what are the key themes and narrative strategies observed in tweets from Russian and Western accounts during the Ukraine conflict? Secondly, what sentiment and thematic patterns of tweets from Russian and Western accounts differ during the Ukraine conflict?

\medskip
One of the challenges in this digital space is distinguishing between genuine information and propaganda \cite{farkas2018disguised}. Propaganda accounts, often associated with state-backed efforts, are known for disseminating emotionally charged content designed to evoke strong reactions and \cite{lan2024exploring} influence public attitudes \cite{freelon2020disinformation}. In contrast, more trusted sources tend to emphasize factual reporting and humanitarian concerns, providing a counterbalance to the misinformation spread by propaganda efforts \cite{lock2019organizational}. Propaganda is closely associated with disinformation \cite{jackson2018distinguishing}, both are often found working side by side \cite{bayer2021disinformation}, as disinformation campaigns are used to promote a particular ideology which in turn becomes propaganda \cite{zhuravskaya2024political}. Understanding the strategies employed by both sides in this information warfare is essential for developing effective countermeasures, especially as it relates to social media platforms. Here, the use of natural language processing and machine learning has become instrumental in analysing large datasets of social media content to identify patterns, themes, and sentiments across different types of accounts \cite{zhou2022understanding}, \cite{antypas2024nlp}, \cite{arowosegbe2023nlp}, \cite{hodorog2022machine}.

\medskip
In our article we employed clustering and classification algorithms. In specific we applied Latent Dirichlet allocation (hereafter LDA) model and a Gaussian Naive Bayes (hereafter GNB) classifier. Other researchers have applied and experimented on similar research \cite{alzanin2022short} using social media data \cite{egger2022topic}. LDA is a probabilistic model and the algorithm generates topics, classifying words among these different topics, according to a given probability distribution \cite{blei2003lda}. We compared this to a GNB which is a machine learning classification technique based on a probabilistic approach that assumes each class follows a normal distribution \cite{huang2018network}. Our results aim to explore the different strategies of information dissemination employed by propaganda and trusted accounts on Twitter/X during the start of the Ukraine conflict. This research not only provides insights into the thematic focus and emotional tone of these narratives but also reveals the presence of coordinated efforts aimed at amplifying specific messages, whether they be hate speech or positive narratives. These types of analyses contribute to a better understanding of the dynamics of information warfare and the role of social media in modern geopolitical conflicts, offering other researchers the methodology and ideas for future research on the influence of digital platforms in shaping public opinion. \cite{rogers2021deplatforming}.

\medskip
We utilised a dataset of tweets from both propaganda accounts and a representative sample of trusted accounts that were systematically scraped using the Twitter API. This dataset is also openly available on Kaggle for other researchers who might be interested, along with data covering a larger time span. In our data, we ranged only from the onset of the conflict in February 2022 until the middle of May 2022. We classified propaganda accounts as those based on their association with state-backed media and a history of spreading disinformation and misinformation, while trusted accounts were selected for their reputation in factual reporting, often verified through independent sources and mainstream media. The dataset included tweets, retweets, likes, replies, and metadata such as timestamps and user information. We did not utilise other features, and only extracted the tweets and retweets and created a dataset of two classes, one for Russian tweets and another for Western tweets. 

\begin{figure}[htbp]
    \centering
    \includegraphics[width=0.6\textwidth]{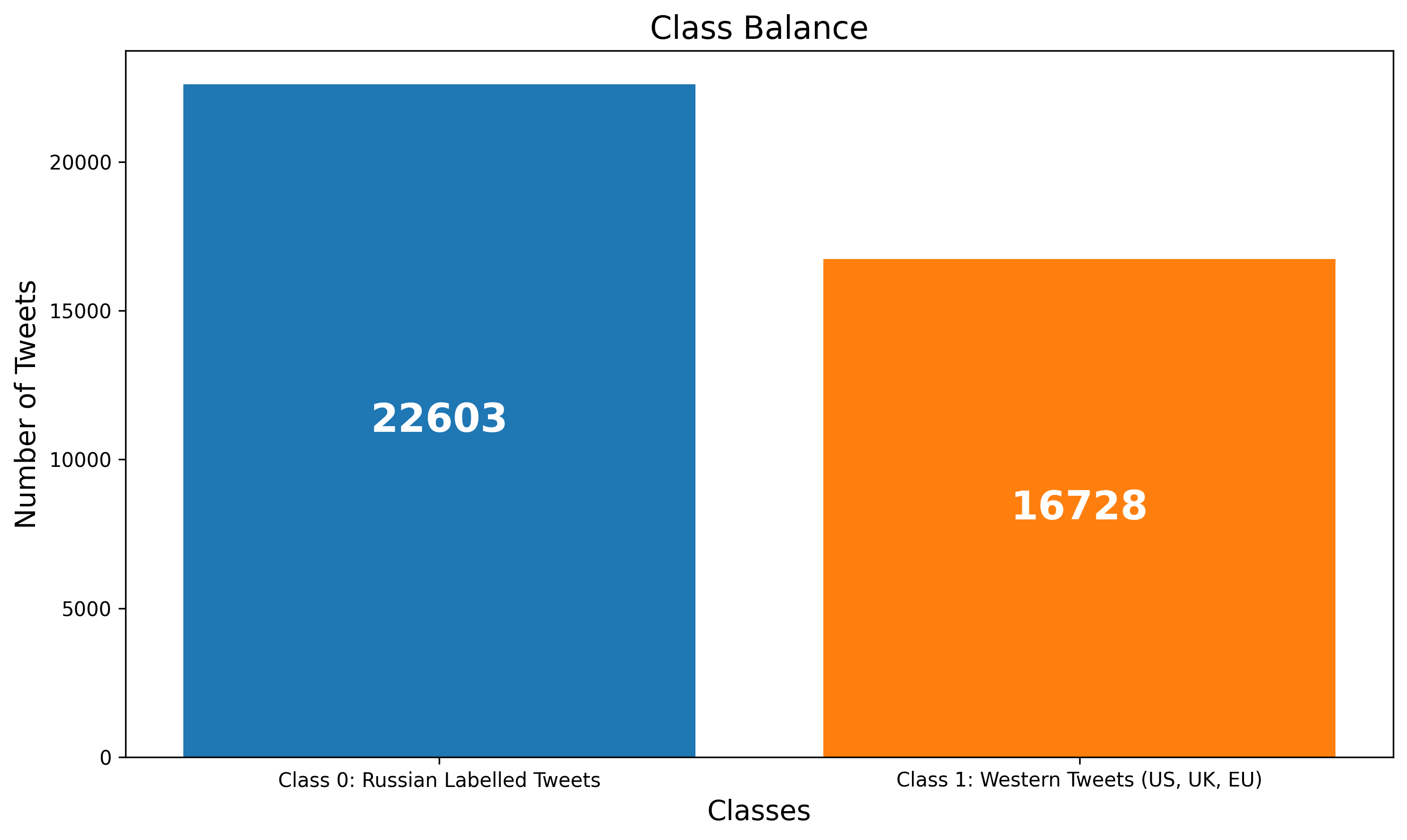}
    \caption{Class Balance of our Dataset}
    \label{fig:classbalance}
\end{figure}

\medskip
Figure \ref{fig:classbalance} shows our class balance, while Table \ref{tab:sampletabletweets} shows a sample of tweets in our dataset which range from hate speech, abusive language, misinformation and disinformation on the Russian class, while the majority on the Western side are factual reporting, event awareness, amplifying atrocities and support for the Ukrainian government.

\begin{table}[ht]     
  \centering
  \footnotesize
  \renewcommand{\arraystretch}{1.15}
  \setlength{\tabcolsep}{4pt}
  \begin{tabularx}{\textwidth}{|Y|Y|}
    \hline
    \textbf{Russian Tweets} \& \textbf{Western Tweets}\\
    \hline
    \textit{Sergey Lavrov: "All these years, our Western colleagues have protected the Ukrainian regime by turning a blind eye to war crimes against civilians. Through their silence, they encouraged the onset of neo-Nazism."} &
    \textit{Volodymyr Zelenskyy: "Russian shelling of a kindergarten in Ukraine that killed at least one child and injured more: What kind of war is that? Were these children neo-Nazi? Or were they NATO soldiers?"}\\
    \hline
    \textit{"The corrupt Ukrainian government is run by fascists. We are criticising Nazis, no problem. Ukraine is full of them and the fascist US is supporting them."} &
    \textit{"Ukraine's sovereignty is supported by a global coalition defending against external aggression. All countries' sovereignty and territorial integrity should be respected."}\\
    \hline
    \textit{"NATO's expansion is a clear act of war. Russia will defend its borders with all necessary force!"} &
    \textit{"NATO continues to provide defensive support to its member states under international law."}\\
    \hline
    \textit{"The so-called 'freedom fighters' in Ukraine are nothing but puppets of the West!"} &
    \textit{"Reports indicate increased humanitarian efforts in Eastern Ukraine to assist those affected by the conflict."}\\
    \hline
  \end{tabularx}
  \caption{Example tweets from Russian and Western accounts (random sample)}
  \label{tab:sampletabletweets}
\end{table}

\subsection{Data Preprocessing}

\begin{figure}[htbp]
    \centering
    \includegraphics[width=0.6\textwidth]{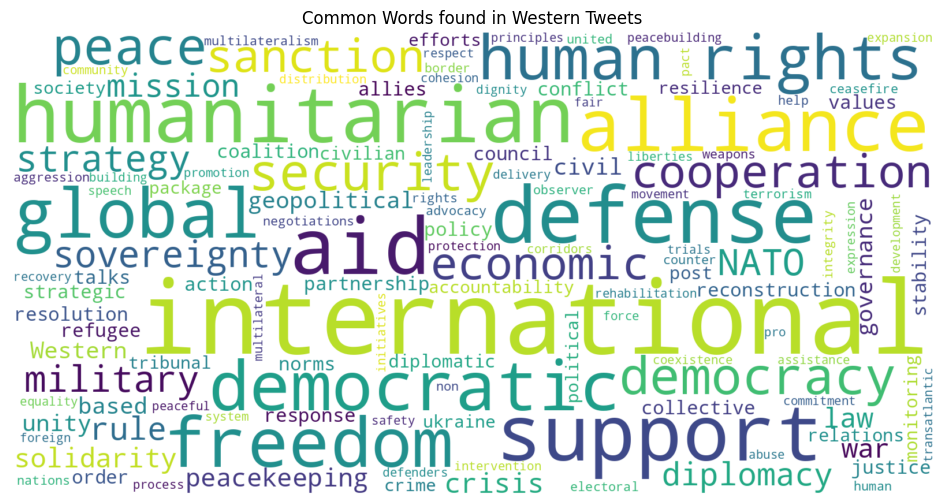}
    \caption{Common Words found in Western Tweets}
    \label{fig:comwordsrus}
\end{figure}

\begin{figure}[htbp]
    \centering
    \includegraphics[width=0.6\textwidth]{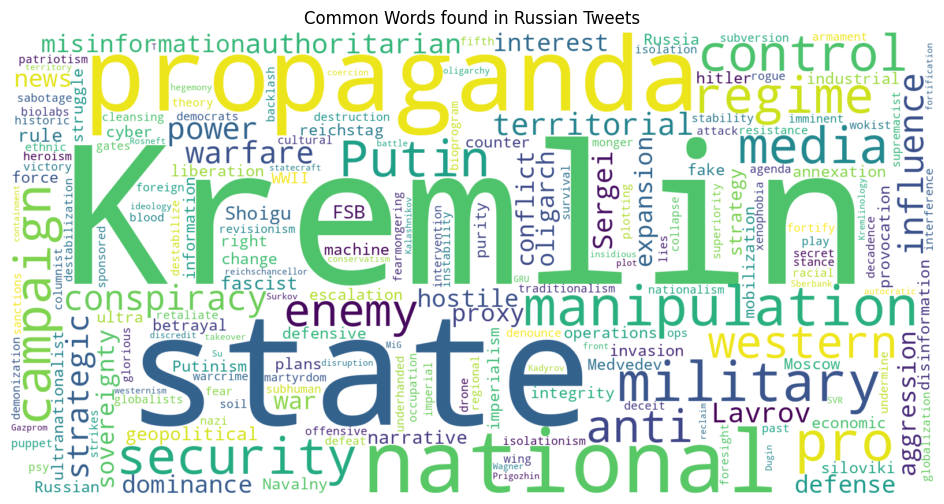}
    \caption{Common Words found in Russian Tweets}
    \label{fig:comwordswest}
\end{figure}

\begin{itemize}[leftmargin=*,align=parleft,nosep]  
  \item \textbf{Tokenisation}: Splitting the text into individual words or
        tokens.  Some common words from both classes can be seen in
        Figure \ref{fig:comwordsrus} for Russian and
        Figure \ref{fig:comwordswest} for Western.
  \item \textbf{Stop-word removal}: Filtering out common words that do not
        contribute to the analysis.
  \item \textbf{Lemmatisation}: Reducing words to their base form
        (e.g.\ “fighting” → “fight”).
  \item \textbf{Noise removal}: Stripping URLs, hashtags, emojis, mentions,
        and non-alphanumeric characters.
\end{itemize}

\subsection{Feature Extraction and Sentiment Analysis}

Following preprocessing, feature extraction was performed using two primary methods:

\begin{itemize}
    \item \textbf{Bag-of-Words (BoW)}: Created a matrix where each word represented a feature, capturing its frequency across the dataset.
    \item \textbf{TF-IDF}: Applied a weighting scheme to adjust word frequency by its importance across the dataset, reducing the impact of commonly used words.
\end{itemize}

Sentiment analysis was conducted GNB, we chose these for its effectiveness in text classification tasks. The classifier, trained on a manually labeled subset of tweets, categorized tweets into positive, negative, or neutral sentiment, allowing us to compare the emotional tone across propaganda and trusted accounts.

\subsection{Thematic Analysis}

To uncover the underlying themes in the tweets, LDA was applied. This generative probabilistic model identifies clusters of words that frequently occur together, interpreted as topics:

\begin{itemize}
    \item \textbf{Defining the number of topics}: Iterative testing was used to determine the optimal number of distinct topics, which we found to be between 5 and 10.
    \item \textbf{LDA on preprocessed text}: This resulted in thematic structures that provided insights into recurring narratives across both propaganda and trusted accounts.
\end{itemize}

\subsection{Clustering Analysis}

Finally, clustering analysis looked at the relationships between accounts based on their tweeting behaviors:

\begin{itemize}
    \item \textbf{Feature Representation}: Each account was represented by a vector comprising tweet content, sentiment scores, and thematic distributions.
    \item \textbf{Clustering}: This algorithm grouped accounts with similar behaviors, and the optimal number of clusters was determined using the Elbow method.
    \item \textbf{Cluster Interpretation}: Analysis revealed patterns of coordinated behavior across Russian accounts, this might suggest potential networks of influence in the information warfare surrounding the Ukraine conflict, although analysing the network effect was not our primarily concern, it was a notable observation. We utilised t-SNE techniques, which reduces the high dimensional data into two dimensional (2D) map.
\end{itemize}

\section{Data and Results}

This section presents the key findings from our analysis of tweets related to the Ukraine conflict. The analysis focuses on thematic exploration, sentiment variations, and narrative strategies.

\subsection{Key Themes and Narrative Strategies}

\begin{figure}[h]
    \centering
    \includegraphics[width=0.6\textwidth]{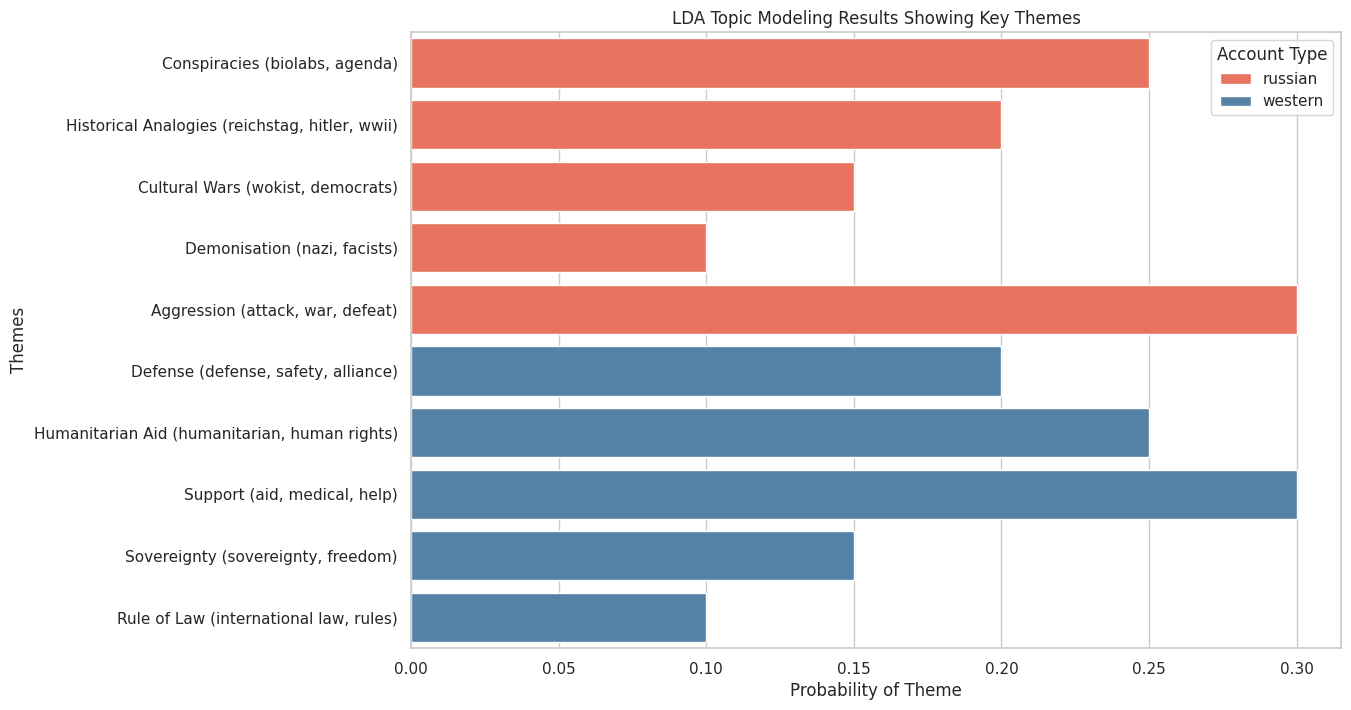}
    \caption{LDA topic modeling results showing key themes.}
    \label{fig:lda_topics}
\end{figure}

\begin{figure}[h]
    \centering
    \includegraphics[width=0.6\textwidth]{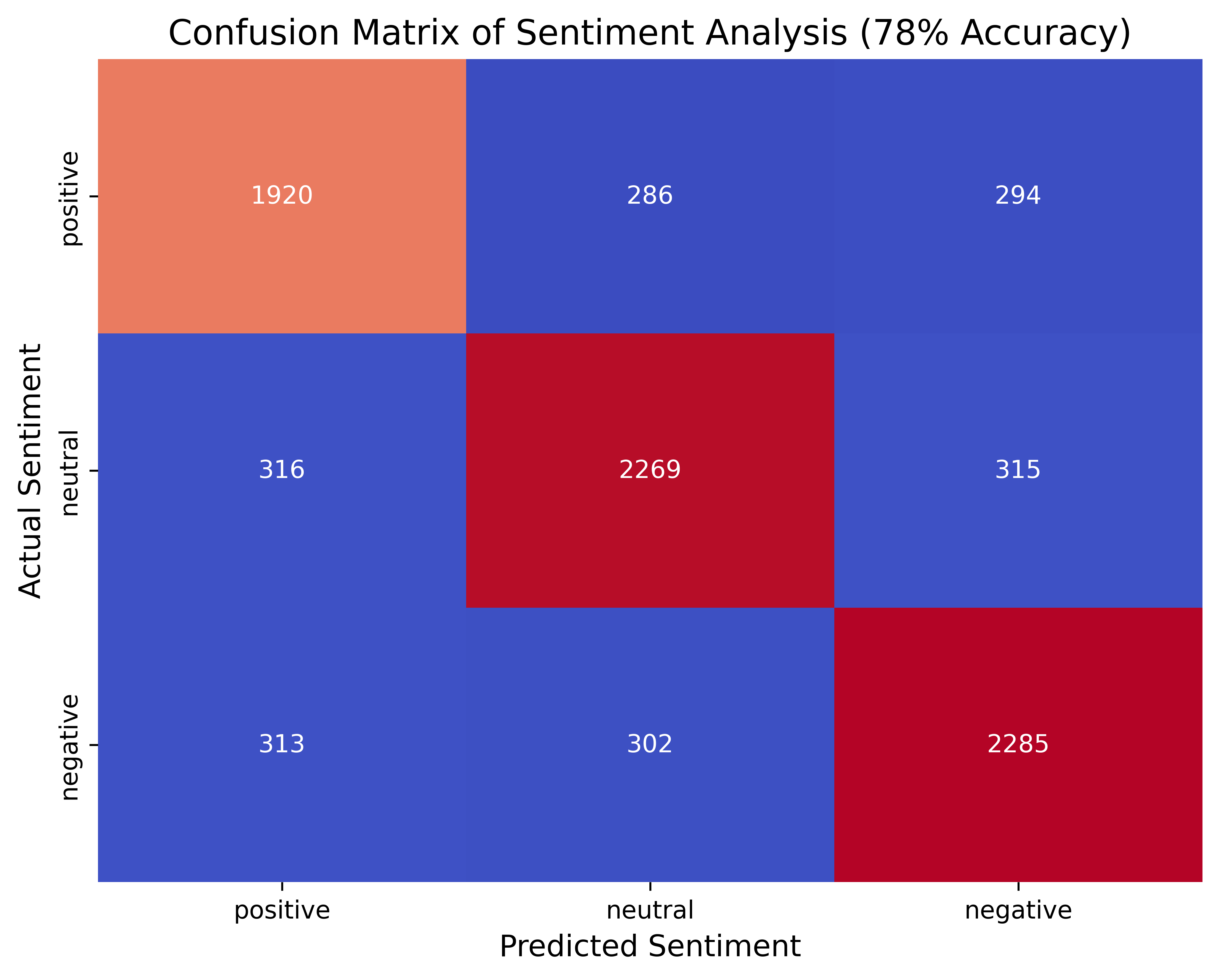}
    \caption{The model has a relatively high accuracy at 78\%, classifying 6400+ correctly out of a total of 8300 predictions.}
\end{figure}

\subsubsection{Western Tweets}

Figure \ref{fig:lda_topics} shows the models results and key themes. Our interpretation here reveals a strong focus on the geopolitical aspects of the Ukraine conflict. Frequently occurring themes include "Ukrainian Support," "Russian Invasion," "Russian Disinformation," "European Alliance," and "NATO," which align with theories of geopolitical narratives, emphasising national security and international order. Keywords such as "Belarus," "Mariupol," and "Kyiv" also highlight the strategic significance of these regions and their role in Ukraine's resistance and sovereignty. The frequent mention of leaders like "Putin" and "Zelensky" reflects the personalisation of the conflict.

\medskip
The reliance on terms like "report," "said," "suggest," "eyewitness," and "video" suggests a strong emphasis on factual reporting, eyewitness accounts, and multimedia content. This approach is characteristic of Western media, which often attempts to maintain credibility and objectivity, despite inherent biases.

\subsubsection{Russian Propaganda Tweets}

\begin{figure}[h]
    \centering
    \includegraphics[width=0.6\textwidth]{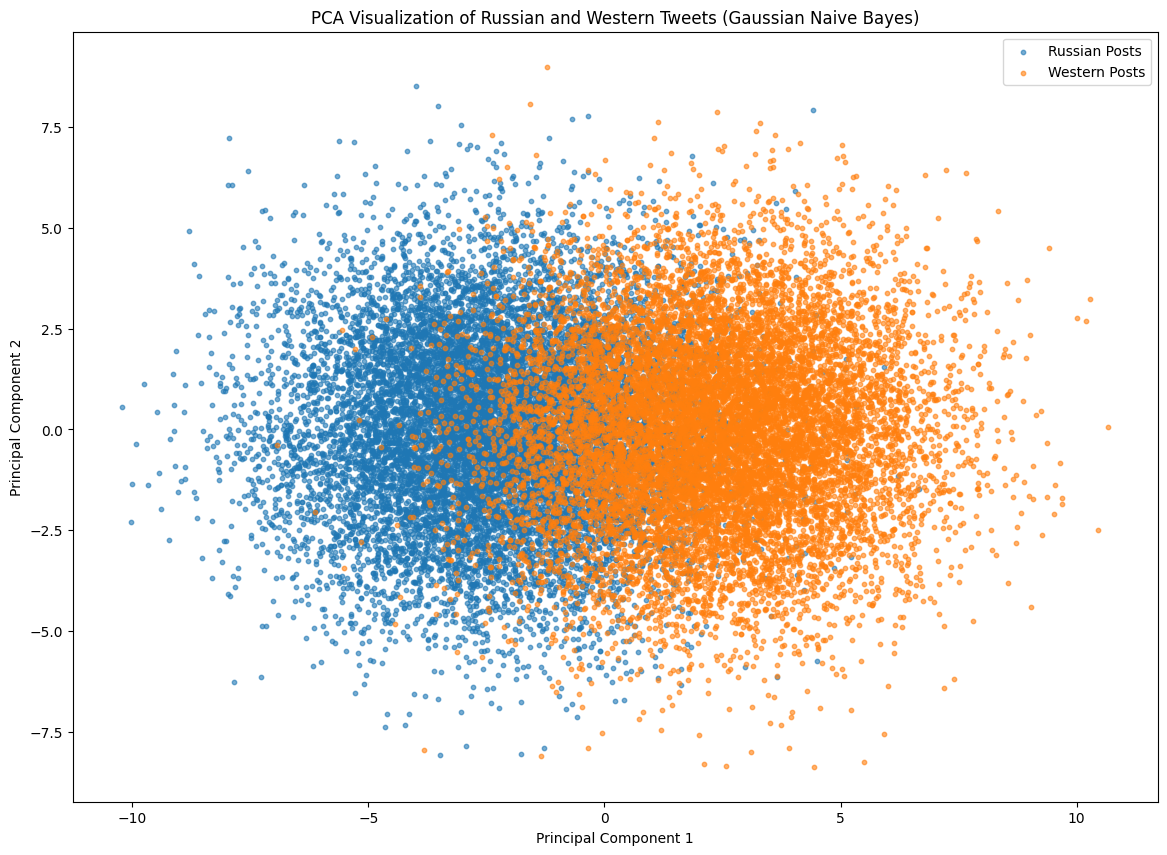}
    \caption{t-SNE of Clusters of GNB}
    \label{fig:clustersgnb}
\end{figure}

In contrast, Russian propaganda tweets show a markedly different set of themes and narrative strategies. The use of German words like "eingriff," "erweiterung," and "reichstag" indicates efforts to influence German-speaking audiences, reflecting a transnational propaganda strategy. The presence of terms such as "biolabs," "reichstag," and "wokist" also points to the use of conspiracy theories and cultural war rhetoric, aiming to evoke fear and justify military actions.

\medskip
The frequent personalisation of the conflict, through mentions of figures like "Joseph Stalin," "Ali Khamenei," and "Adolf Hitler," is designed to humanise the geopolitical struggle. This approach, known in propaganda studies as the "great man" technique, leverages the symbolic power of leadership figures to influence public opinion \cite{jowett2019propaganda}. Additionally, the use of derogatory language like "beta," "simp," and "autist" indicates a strategy of trolling and memes in attempt to polarise and degrade people, which is well-documented in studies of online radicalisation.

\begin{figure}[h]
    \centering
    \includegraphics[width=0.6\textwidth]{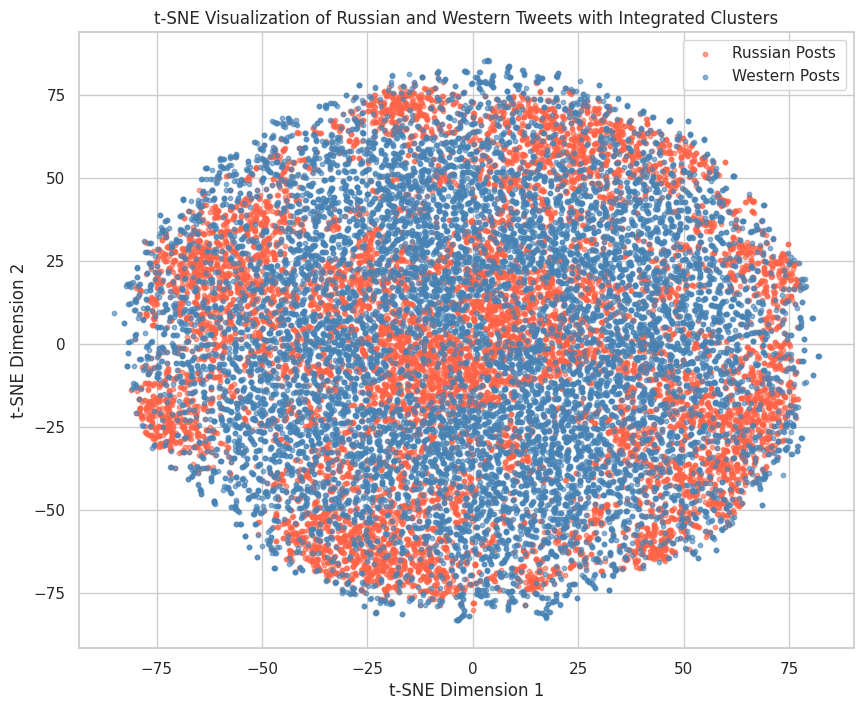}
    \caption{t-SNE of Clusters of LDA Model}
    \label{fig:western_themes}
\end{figure}

\subsection{Sentiment Analysis and Narrative Strategies}

Sentiment analysis revealed substantial differences between Russian and Western tweets. Russian propaganda is characterised by a predominantly negative sentiment, using emotionally charged language to provoke fear and justify aggressive actions. Terms like "Nazi," "attack," and "force" frequently appeared, supporting a narrative of existential threat \cite{ojala2019role,jowett2019propaganda}. In contrast, Western tweets exhibited more varied sentiment, often leaning towards neutral or positive tones, especially when discussing humanitarian efforts. This reflects a narrative strategy focused on factual reporting and the ethical obligations to support Ukraine.

\begin{figure}[h]
    \centering
    \includegraphics[width=0.6\textwidth]{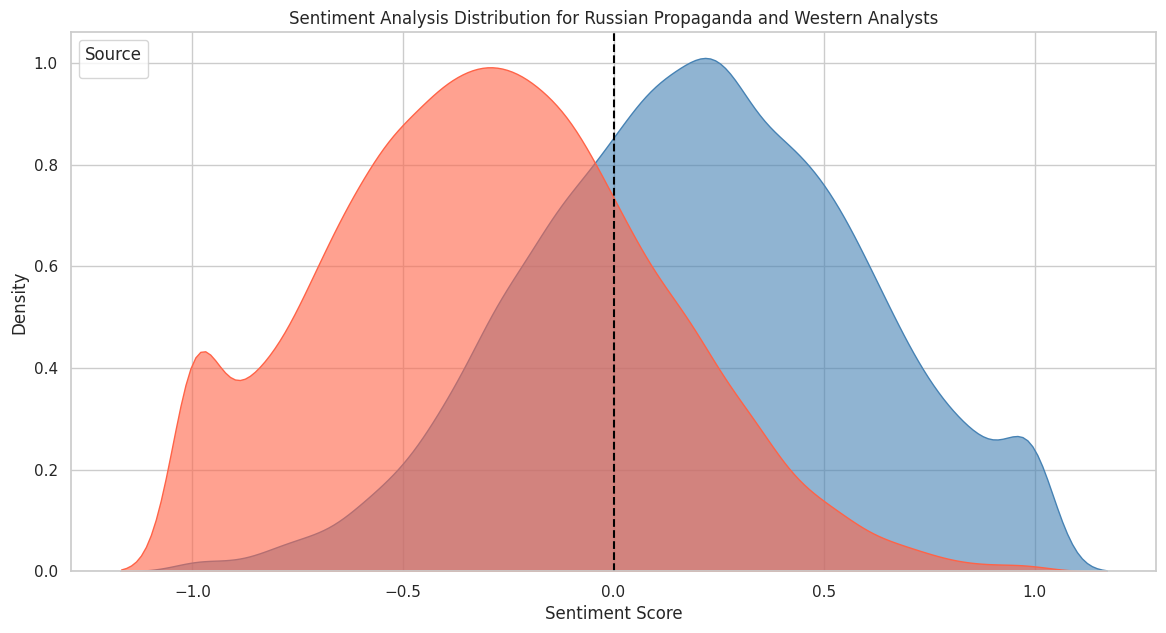}
    \caption{Sentiment analysis results across different accounts.}
    \label{fig:sentiment_analysis}
\end{figure}

\medskip
Our findings show the stark differences in how the conflict is portrayed on Twitter/X, demonstrating the platform's role as an active battleground for information warfare, where narratives are constructed, contested, and propagated.

\section{Conclusion}

This research article provided an analysis of the narrative strategies used by Russian propaganda accounts and Western accounts on Twitter at the start of the Ukraine conflict. The findings reveal contrasts in the dissemination of information, with Russian accounts often relying on emotionally charged language, conspiracy theories, the spread of misinformation/disinformation and historical analogies to evoke fear and justify military actions. In contrast, Western analysts tended to focus on factual reporting and humanitarian concerns.

\medskip
We also acknowledge that due to the limited time range, this does not provide an accurate representation of what narratives are in the overall context of the conflict and that representation through Tweets and machine learning models has limitations in so far. Our model accuracy at 78\% also suggests that a large enough sample is misclassified and misunderstood. However, overall we these differences suggest to us that the role of social media is an important battleground for information warfare, as the consequences of this information often leads to radicalisation or influence of public opinion which can have an effect on the support of the those involved in the conflict. Narrative strategies are continually adapted to the evolving context of the conflict and the importance of media literacy to identify propaganda and misinformation/disinformation cannot be understated. There is also the need for ongoing research into the mechanisms of digital propaganda, particularly as these tactics continue to influence public perception on social media.

\bibliographystyle{plainurl} 
\bibliography{references}  

\end{document}